\documentclass[letter]{article}
\usepackage{INTERSPEECH2021}
\usepackage{amsmath,graphicx}
\usepackage{booktabs}
\usepackage{multirow}
\usepackage[normalem]{ulem}
\useunder{\uline}{\ul}{}
\usepackage{url}

\title{Improving low-resource ASR performance with untranscribed out-of-domain data}
\name{Jayadev Billa\thanks{
This research is based upon work supported in part by the Office of the Director of National Intelligence (ODNI), Intelligence Advanced Research Projects Activity (IARPA), via contract \# FA8650-17-C-9116. The views and conclusions contained herein are those of the authors and should not be interpreted as necessarily representing the official policies, either expressed or implied, of ODNI, IARPA, or the U.S. Government. The U.S. Government is authorized to reproduce and distribute reprints for governmental purposes notwithstanding any copyright annotation therein.}}

\address{Information Sciences Institute,\\
University of Southern California, Marina del Rey, CA 90292, USA}
\email{jbilla@isi.edu}
\begin{document}

\maketitle
\begin{abstract}
Semi-supervised training (SST) is a common approach to leverage untranscribed/unlabeled speech data to improve automatic speech recognition performance in low-resource languages. However, if the available unlabeled speech is mismatched to the target domain, SST is not as effective, and in many cases performs worse than the original system. In this paper, we address the issue of low-resource ASR when only untranscribed out-of-domain speech data is readily available in the target language. Specifically, we look to improve performance on conversational/telephony speech (target domain) using web resources, in particular YouTube data, which more closely resembles news/topical broadcast data. Leveraging SST, we show that while in some cases simply pooling the out-of-domain data with the training data lowers word error rate (WER), in all cases, we see improvements if we train first with the out-of-domain data and then fine-tune the resulting model with the original training data. Using 2000 hours of speed perturbed YouTube audio in each target language, with semi-supervised transcripts, we show improvements on multiple languages/data sets, of up to 16.3\% relative improvement in WER over the baseline systems and up to 7.4\% relative improvement in WER over a system that simply pools the out-of-domain data with the training data.
\end{abstract}

\section{Introduction}
\label{sec:intro}

Speech recognition systems rely on the availability of large amounts of labeled, i.e. transcribed speech data, preferentially data that is similar to the target environment in which these systems will be deployed, for optimal performance. Unfortunately for domains outside of broadcast media, collection of speech data is an involved process, e.g. if the target domain is conversational/telephony speech, it would involve a targeted data collection effort with the informed consent of 100s to 1000s of participants. Moreover, labeling/transcription of speech data at scale is both costly and time consuming. Given this state of affairs, it would be useful to find an approach that can bootstrap from a limited in-domain training set and use untranscribed out-of-domain speech data, in particular broadcast media and user generated content on YouTube, to improve performance in the target domain. In this paper, we combine semi-supervised training with transfer learning to demonstrate one approach to accomplish this goal.

Semi-supervised training (SST), an approach to use untranscribed speech to train ASR models, started with early efforts to harvest training data from speech transcribed with a bootstrap ASR system, followed by the selection of a subset of transcripts based on a threshold applied to the confidence in the truth of the transcript words~\cite{DBLP:conf/interspeech/ZavaliagkosSCB98}. This harvested data is then combined with existing labeled data to create a larger training set with which to build better performing models. Later efforts applied this approach to ever larger data collections, e.g. \cite{DBLP:conf/icassp/MaMKS06} and most recently to neural network based ASR systems, e.g. \cite{DBLP:conf/icassp/MaM07,DBLP:conf/asru/VeselyHB13,DBLP:conf/icassp/ThomasSCH13,DBLP:conf/icassp/ManoharHPK18,DBLP:journals/speech/YuGWW10}. 

Transfer learning is the broad approach of using an existing trained model, or elements thereof, for a new task or domain; see \cite{pan2009survey} for a general treatment of transfer learning, \cite{bengio2012deep} for a deep learning perspective, and \cite{DBLP:conf/slt/SwietojanskiGR12,DBLP:conf/icassp/HuangLYDG13,ghahremani2017} for a speech recognition modeling perspective. Fine-tuning, in the context of neural models, is a transfer learning approach that involves retraining an existing neural model for a new task/domain with a lower learning rate, with data from the target task.

Semi-supervised training has been an integral tool in our work on the IARPA MATERIAL~\cite{material17} program. Briefly, the IARPA MATERIAL program goal is to allow English queries and information extraction across text and audio sources in low-resource non-English languages. A key element of information extraction from audio sources is the generation of transcripts for the audio via ASR. Separate from the low-resource aspect of the ASR training data, an additional complicating factor is that  test audio documents are derived from a range of domains, with the majority taken from broadcast media or similar sources. In this particular scenario, we find that the availability of large amounts of broadcast media and user generated content, sourced from YouTube, allows us to overcome the lack of broadcast media like training data with extensive use of SST on the YouTube data. However, on conversational speech, due to the domain mismatch, SST with YouTube data typically does not help and in many cases hurts performance - leading to the problem scenario we describe above. 

Our approach to addressing this problem is to first build a bootstrap model with the limited labeled in-domain data and then use that initial model with SST to build an out-of-domain ASR model. The traditional approach to SST involves pooling the labeled data with the high-confidence transcripts/audio from the unlabeled set -- high-confidence meaning that we use an appropriate metric to select the best data to train on. In our work, we simply use a median threshold on the per frame likelihood as the selection criteria. As an alternative, we can omit the pooling stage and simply train on the newly labeled out-of-domain data exclusively.
Regardless of how we build the out-of-domain model, we can then fine-tune, essentially retraining with a lower learning rate, this model with the original in-domain training data. We find that this approach, across a variety of data sets, improves upon the original bootstrap model performance and, at least in the languages and data sets evaluated, consistently outperforms a model trained with pooling the original data with out-of-domain data in an SST framework.

In Section~\ref{swbd-expt} we investigate this approach with a subset of the Switchboard (SWBD) corpus~\cite{swbd} as our in-domain training set, a subset of the Fisher corpus~\cite{DBLP:conf/lrec/CieriMW04} as a control in-domain untranscribed set, and YouTube data to illustrate potential improvements. Section~\ref{mat-expt}
details experiments on several MATERIAL languages using the program provided labeled data as well as data from YouTube. Finally, in Section~\ref{expt-disc}, we conclude with a summary and implications of this work.

\section{Switchboard experiments}
\label{swbd-expt}

\subsection{Data sets and experimental setup}
\label{ssec:swbd-sets}

In our initial experiments, we will use a 150 hour subset of speed perturbed SWBD corpus (nominally 50 hours of original data), as our low-resource in-domain training set. Likewise, we select 2000 hours of speed perturbed Fisher corpus data (nominally 667 hours of Fisher data) as an in-domain control set. With Fisher, we experiment 
with both the original transcripts as well as transcripts generated by decoding the data with the bootstrap SWBD ASR model. For our out-of-domain data, we use a random selection of English broadcast media data available on YouTube, again 2000 hours after speed perturbation, and generate transcripts with the bootstrap SWBD model. 

For evaluation, we use the HUB5 Eval2000 data set (LDC2002S09/LDC2002T43) which consists of both SWBD and CallHome subsets. The SWBD subset is similar to the training data, telephone conversations on a prescribed set of topics between individuals unfamiliar with one another, whereas CallHome, while still being telephone conversations, is mismatched in that they are conversations between family or close friends unconstrained by topic. From this perspective, CallHome may provide a better correspondence with performance improvements in real life. In these experiments, we constrain the language model to a trigram language model (LM) trained on the complete set of SWBD transcripts.

For ASR system development, we use the Kaldi toolkit~\cite{PoveyASRU2011}, using the end-to-end LF-MMI approach ~\cite{DBLP:conf/interspeech/HadianSPK18} to building acoustic models using 40-dimensional MFCC features on speed perturbed~\cite{KoPPK15} data. YouTube data is downsampled to 8kHz, to match the SWBD and Fisher data, before feature generation. In terms of the actual training process, we use Kaldi's existing SWBD and Fisher recipes\footnote{\scriptsize available at \texttt{https://github.com/kaldi-asr/kaldi} in \texttt{egs/swbd/s5c} and \texttt{egs/fisher\_swbd/s5}} 
to preprocess the Switchboard and Fisher data respectively for ASR model training. The trigram language model is also generated by the process detailed in the same SWBD recipe. As our canonical acoustic model architecture, we use a 15-layer TDNN-F~\cite{DBLP:conf/interspeech/PoveyCWLXYK18} model architecture in all experiments in this paper. Lastly, with regards to training with end-to-end training LF-MMI, we use this approach for no other reason beyond that it offers a more streamlined process for experimentation. 

\subsection{Results}
\label{ssec:swbd-res}

Table~\ref{tab:swbd-res} summarizes the various experiments. We see the out-of-domain YouTube data trained model underperforms the baseline system due to the mismatched domain; adding in SWBD data and training on the pooled data improves performance, though on SWBD, performance lags its baseline, and CallHome performance is essentially on par with its baseline performance. However, if we fine-tune the under-performing YouTube data or pooled data trained model, we eliminate the performance gap on SWBD. On CallHome, we observe a 6\% relative improvement in WER compared to the baseline with both fine-tuned models. Furthermore, we note that performance after fine-tuning of both the pooled and YouTube only trained models is essentially the same.

Using Fisher, an "in-domain" data set, with original transcripts, we no longer see the degradation we observed with the YouTube data alone, and pooling the Fisher data with SWBD further improves performance. Similar to what we observed with the YouTube experiments, if the Fisher model is fine-tuned we see a further improvement. If we switch to using transcripts generated with the bootstrap SWBD model, apart from noting that performance lags that of models trained with true transcripts, we observe the same trends.

From these results, it appears that fine-tuning models, trained on out-of-domain data, can recover from domain mismatch and improve over both the baseline and models trained on the pooled data set.
\begin{table}[ht]
\centering
\caption{Comparison of WER improvements on the Eval2000 test set for Switchboard (SWBD) and CallHome (CH) subsets, using models trained on different variations of SWBD/YouTube (YT) and SWBD/Fisher data. Fisher data subset using original transcripts and semi-supervised transcripts are distinguished with OT/SST respectively. Pooled implies the model is trained with combined data, fine-tuning always uses the SWBD data. \%RI refers to relative improvement over the corresponding baseline.}
\label{tab:swbd-res}
\begin{tabular}{@{}lcccc@{}}
\toprule
\multirow{2}{*}{Training} & \multicolumn{2}{c}{WER}                     & \multicolumn{2}{c}{\%RI}           \\ \cmidrule(l){2-5} 
                       & SWBD & CH   & SWBD  & CH   \\ \midrule
SWBD (baseline)        & 12.5 & 24.8 & -     & -    \\ \midrule
YouTube                & 15.5 & 26.9 & -24.0 & -8.5 \\
\textit{+ fine-tuning} & 12.5 & 23.3 & 0.0   & 6.0  \\ \cmidrule(l){2-5} 
Pooled YouTube/SWBD    & 13.8 & 24.7 & -10.4 & 0.4  \\
\textit{+ fine-tuning} & 12.3 & 23.5 & 1.6   & 5.2  \\ \midrule
{\ul Fisher w/ OT}        & \multicolumn{1}{l}{} & \multicolumn{1}{l}{} & \multicolumn{1}{l}{} & \multicolumn{1}{l}{} \\
Fisher                 & 12.5 & 22.2 & 0.0   & 10.5 \\
\textit{+ fine-tuning} & 11.3 & 21.3 & 9.6   & 14.1 \\ \cmidrule(l){2-5} 
Pooled Fisher/SWBD     & 11.2 & 20.5 & 10.4  & 17.3 \\
\textit{+ fine-tuning} & 10.8 & 20.7 & 13.6  & 16.5 \\ \midrule
{\ul Fisher w/ SST}       & \multicolumn{1}{l}{} & \multicolumn{1}{l}{} & \multicolumn{1}{l}{} & \multicolumn{1}{l}{} \\
Fisher                 & 12.4 & 22.8 & 0.8   & 8.1  \\
\textit{+ fine-tuning} & 11.6 & 22.1 & 7.2   & 10.9 \\ \cmidrule(l){2-5} 
Pooled Fisher/SWBD     & 12.1 & 22.6 & 3.2   & 8.9  \\
\textit{+ fine-tuning} & 11.5 & 22.3 & 8.0   & 10.1 \\ \bottomrule
\end{tabular}
\end{table}

\section{MATERIAL experiments}
\label{mat-expt}

\subsection{Data sets and experimental setup}
\label{ssec:mat-sets}

The SWBD experiments in Section~\ref{swbd-expt} allowed us to compare and contrast performance with an in-domain data set. In reality, if we have in-domain data, we would train directly on it. The MATERIAL program offers an environment to test our approach in a realistic use case: a limited in-domain labeled training set, and otherwise unconstrained in terms of usage of audio/text resources obtained on the web. In particular, we will investigate performance on three MATERIAL program languages: Farsi, Kazakh, and Lithuanian. For the ASR component of the program, a small training set, the \textbf{build} set, consisting of approximately 40 hours of mostly telephony speech is provided, as well as a mixed speech (telephone, broadcast news, and conversation) test set, the \textbf{analysis} set. In this set of experiments, since we are focused on the in-domain performance only, all results presented here pertain to the telephone/conversational speech subset of the \textbf{analysis} set. Table~\ref{tab:mat-data} summarizes both acoustic modeling and language modeling data. For each language, language modeling data is largely web crawl data, but we benefit from earlier work in addition to our web crawl efforts. Concretely, for Farsi, we use a subset of MirasText~\cite{DBLP:conf/lrec/SabetiFCNV18}\footnote{details at \texttt{\scriptsize https://github.com/miras-tech/MirasText}}, and for Kazakh we additionally use a subset of the WMT19 Kazakh data. 
\begin{table}[ht]
\centering
\caption{Data used in MATERIAL language experiments. Training data corresponds to each language's build data set provided as part of MATERIAL program. CS is conversational/telephone speech, WB (wide-band) groups the remaining non-telephone speech in the respective build set.}
\label{tab:mat-data}
\begin{tabular}{@{}lccc@{}}
\toprule
\multirow{2}{*}{Language} & \multicolumn{2}{c}{Training data (hr)} & \multirow{2}{*}{LM data  (\#words)} \\ \cmidrule(lr){2-3}
           & CS   & WB  &      \\ \midrule
Farsi      & 36.3 & -   & 1B   \\
Kazakh     & 43.0 & 6.5 & 120M \\
Lithuanian & 46.3 & 7.0 & 170M \\ \bottomrule
\end{tabular}
\end{table}

As our out-of-domain data, we first download audio from YouTube in each language and then decode with available in-language ASR models. The ASR models used in decoding the YouTube data were originally bootstrapped with the build data but were subsequently retrained with YouTube data using SST. Once transcripts for a 2000 hour subset of speed perturbed data are generated, we combine YouTube audio and generated transcripts to create an out-of-domain training set for each language. Once again the 2000 hour YouTube data for each language nominally represents 667 hours of original YouTube audio but is not speed perturbation of a specific 667 hour selection of YouTube audio, since transcripts are generated on and selected from speed perturbed data.

The ASR training setup remains as before, we use the same acoustic model architecture with end-to-end LF-MMI training. For language modeling, given the multiple sources of data, we use language model interpolation to combine source specific trigram LMs using the SRILM Toolkit~\cite{DBLP:conf/interspeech/Stolcke02} using weights that minimize the perplexity on the conversational speech subset of the respective language's analysis set. For any specific language, the LM is fixed across all experiments.

\subsection{Results}
\label{ssec:mat-res}

 Table~\ref{tab:mat-res} details results across all three languages. As observed on the SWBD experiments, we see again that the out-of-domain YouTube data by itself always hurts performance. Similarly, training on YouTube and build data always outperforms models trained only with the YouTube data, however, the improvement is not guaranteed to improve on the build only models -- on Farsi, we see a 4\% relative increase in WER with the pooled data trained model, but both Kazakh and Lithuanian benefit from the pooled data, with 12.2\% relative and 11\% relative decrease in WER over their respective baselines. Across all three languages, we see a consistent improvement in WER with fine-tuning. Again we observe that regardless of whether we fine-tune the YouTube only trained model or the pooled data trained model, the resulting models  perform about the same. 

Another general method used in low-resource languages is to apply transfer learning on a model in another language by replacing the output layers of the acoustic model with randomly initialized layers of the appropriate size for the target language. To compare our fine-tuning approach against this method, we take an existing GALE Arabic model trained on an 800 hour subset of GALE Arabic Phase 2-4 datasets (available from LDC) consisting of broadcast news and broadcast conversation in Arabic and retrain using the build data for each language. Table~\ref{tab:mat-res} also details the results of these transfer training experiments. In general, we see the fine-tuning approach compares favorably with transfer models; on Farsi and Kazakh the transfer model approach is slightly worse than fine-tuning, on Lithuanian, the transfer model performs better.
\begin{table}[ht]
\centering
\caption{Comparison of WER improvements on CS subset of the analysis set for MATERIAL languages. Fine-tuning is always with the language's respective build set.}
\label{tab:mat-res}
\begin{tabular}{@{}llcc@{}}
\toprule
Language   & Training               & \%WER & \%RI \\ \midrule
Farsi      & Build (baseline)       & 45.4  & -             \\ \cmidrule(l){2-4} 
           & YouTube                & 60.4  & -33.0         \\
           & \textit{+ fine-tuning} & 43.9  & 3.3           \\ \cmidrule(l){2-4} 
           & Pooled YouTube/Build   & 47.2  & -4.0          \\
           & \textit{+ fine-tuning} & 43.7  & 3.7           \\ \cmidrule(l){2-4}
           & Arabic Transfer        & 44.3  & 2.4           \\ \midrule
Kazakh     & Build (baseline)       & 55.1  & -             \\ \cmidrule(l){2-4} 
           & YouTube                & 68.8  & -24.9         \\
           & \textit{+ fine-tuning} & 46.1  & 16.3          \\ \cmidrule(l){2-4} 
           & Pooled YouTube/Build   & 48.4  & 12.2          \\
           & \textit{+ fine-tuning} & 46.4  & 15.8          \\ \cmidrule(l){2-4}
           & Arabic Transfer        & 47.0  & 14.7          \\ \midrule
Lithuanian & Build (baseline)       & 48.9  & -             \\ \cmidrule(l){2-4} 
           & YouTube                & 63.4  & -29.7         \\
           & \textit{+ fine-tuning} & 42.9  & 12.3          \\ \cmidrule(l){2-4} 
           & Pooled YouTube/Build   & 43.5  & 11.0          \\
           & \textit{+ fine-tuning} & 42.8  & 12.5          \\ \cmidrule(l){2-4}
           & Arabic Transfer        & 41.3  & 15.5          \\ \bottomrule
\end{tabular}
\end{table}

\section{Discussion}
\label{expt-disc} 

In this paper we demonstrate, over multiple languages and data sets, the utility of combining SST and fine-tuning to improve performance in a particular domain when we have limited in-domain data and access only to out-of-domain YouTube data. We find that regardless of the choice of model to fine-tune, i.e. model trained on out-of-domain data only or pooled with the in-domain data, we always improve upon the baseline model. Depending on the context, one model may be more relevant/efficacious than the other, e.g. if pooled data trained models are readily available, as in our case, since we use incremental SST~\cite{DBLP:conf/icassp/KhonglahMDBMB20}, it is more efficient to fine-tune the existing pooled data trained model. Alternatively, if training from scratch, it might be quicker to train the model on out-of-domain data only before fine-tuning, or if ensembling the output of multiple systems, it would make sense to train and fine-tune both models. Lastly, we show that transfer modeling and our fine-tuning approach provide similar performance improvements, offering another technique that can be applied to low-resource ASR system development.



\bibliographystyle{IEEEtran}

\bibliography{references}

\end{document}